# Zero-error dissimilarity based classifiers[1]

**Robert P.W. Duin and Elżbieta Pękalska**

Delft Pattern Recognition Group
Faculty of Applied Sciences
Lorentzweg 1, 2628 CJ Delft, The Netherlands
r.duin@ieee.org

## Abstract

We consider general non-Euclidean distance measures between real world objects that need to be classified. It is assumed that objects are represented by distances to other objects only. Conditions for zero-error dissimilarity based classifiers are derived. Additional conditions are given under which the zero-error decision boundary is a continues function of the distances to a finite set of training samples. These conditions affect the objects as well as the distance measure used. It is argued that they can be met in practice.

## 1 Introduction

In pattern recognition the real world objects like characters or faces are traditionally represented in a feature space. Generalisation is performed by training classifiers in such a space using a set of examples, the training set. In some applications like shape recognition, good features may be difficult to find and dissimilarities between objects are used for representation [1]. Usually the nearest neighbour rule is then used as a classifier.

In our previous work we showed that instead of the nearest neighbour rule also other, traditional classifiers like Fisher's Linear Discriminant may be used for generalisation. To that purpose we studied the use of dissimilarity spaces as well as Euclidean embedding for representing the dissimilarities between the objects in the training set as well as between the training objects and test objects to be classified. We found that such dissimilarity based classifiers may perform better and may be evaluated faster than the nearest neighbour rule.

A problem that had to be faced in building dissimilarity based spaces on distances measures practically used in pattern recognition is that such distance measures are frequently not Euclidean embeddable and are even sometimes not metric. Nevertheless, they may perform well and it remains of practical interest to study their properties fundamentally.

An intriguing aspect of the dissimilarity based approach to pattern recognition is that the distance measures of interest, although not Euclidean, may be given other interesting properties. One of them is that they may built a so-called Hausdorff space in which distances are zero if and only if the objects are identical. In case the real world objects can be unambiguously labelled, which is rather natural for real world objects, this opens the road to build so-called zero-error classifiers. Such classifiers make use of the non-overlap of the classes and define the classification function in the gap between them.

---

1. The original paper was written in 2003 when the first author was visiting Anil Jain, MSU in East Lansing, USA. The paper was not accepted for publication as it 'tried to proof an obvious truth'. Nevertheless it was for the authors a significant step as it made them clear that there should be in some way a continuous relation between the real world objects and their representation to have the prospect of a zero-error classifier defined by a finite training set.

For non-Euclidean embeddable datasets, however, we don't have yet a space in which this may be performed. For such datasets the problem of constructing such classifiers is not yet solved. In this paper we will approach it fundamentally and start by defining a topology based on a neighbourhood system defined on a given dissimilarity measure. We will define conditions for such measures and for the class of objects under consideration and proof two fundamental theorems based on these conditions.

First, it will be proved that under very general conditions for the dissimilarity measure, including non-metricness, a zero error classifier exists.

Second, general conditions with respect to the class of objects will be given under which such a classifier can be based on a finite set of objects.

Together, these two theorems show that a zero-error classifier may be constructed in practice using a finite training set.

## 2 Intuition

Before giving a formal proof we will give some intuition on what we are aiming at.

Consider a two-class dissimilarity based classification problem, e.g. face recognition from 2D images. Let for each of the classes the images of the two persons A and B be sampled from a set for which parameters describing light circumstances, position, face expression, etcetera, vary continuously. Let us consider a dissimilarity measure $D(x,y)$ between face images $x$ and $y$ that is continuous in the parameters generating $x$ and $y$. Let $D(x,y) = 0$ iff $x = y$.

Now the infinite set of images of A as well as the infinite set of images of B constitute under some conditions for $D(\bullet)$ both connected sets $A_D$, respectively $B_D$, defined on $D(\bullet)$. For all images of A there is a one to one relation with an element in $A_D$. Similar for B and $B_D$. Very similar images are close in such sets. Between any two images of the same class exists a connected path between the two elements representing the two images.

For classification problems like the given one it can be assumed that no two images, one from A and one from B have a dissimilarity $D < \delta$. Consequently the two sets $A_D$ and $B_D$ do not intersect and are thereby entirely separable.

## 3 Conditions

We will proof the above formally if the following conditions hold for dissimilarity measure $D(x,y)$, $\forall x,y \in X$, in which X is the entire set of objects of interest.

**Condition 3.1** $D(x,y) \in \Re$, $0 \leq D(x,y) < \infty$.

**Condition 3.2** $D(x,y)$ is continuous in the parameters that generate the objects $x$ and $y$.

**Condition 3.3** $D(x,y) = 0$ iff $x = y$.

**Condition 3.4** $D(x,y) > \delta$, for some $\delta$, $0 < \delta < \infty$ iff $(x \in A \land y \in B) \lor (x \in B \land y \in A)$.

**Lemma 3.1** For a sufficiently small $\varepsilon$, $\forall x \in A \; \exists y \in A \; D(x,y) < \varepsilon \; y \neq x$, similar for $x \in B \; y \in B$.
This directly follows from the conditions 3.2, 3.3 and 3.4.

## 4 The existence of a zero-error classifier

In this section we will proof that there exists a classifier that under the given conditions all $x$ correctly classifies. First we define a neighbourhood basis for $x$ and on top of that neighbourhoods and neighbourhood systems.

**Definition 4.1** The neighbourhood basis of $x$, NB($x$) is the set of all $y$ with
D($x,y$) < $\varepsilon$, 0 < $\varepsilon$ < $\delta$.

**Definition 4.2** If P(X) is the power set of X, then N $\in$ P(X) is a neighbourhood of $x$ if
NB($x$)$\subseteq$N

**Definition 4.3** A neighbourhood system $N(x)$ is the collection of all neighbourhoods of $x$.

As a result has each $x$ in A in all its neighbourhoods some $y$ in A:

**Lemma 4.1** $\forall x \in$ A $\exists y \in$ N $\forall$N$\in N(x)$ $y \neq x$ $y \in$ A.
From definition 4.1 it follows that $\exists y \in$ A with D($x,y$) < $\varepsilon$ is equivalent with $\exists y \in$ NB($x$), which is equivalent with $\exists y \in$ N $\forall$N$\in N(x)$ as $\forall$N$\in N(x)$ NB($x$)$\subseteq$N (definitions 4.2 and 4.3). Substitution in lemma 3.1 shows the result.

The neighbourhood basis of all x in A contains no points of B

**Lemma 4.2** $\forall x \in$ A $\forall y \in$ B $y \notin$ NB($x$) $x \notin$ NB($y$).
This follows from the condition 3.4 on the dissimilarities between objects in A and B and definition 4.1 for NB($x$).

So all $x$ in A have a neighbourhood that contains no points of B

**Lemma 4.3** $\forall x \in$ A $\exists$N$\in N(x)$ N$\cap$B=$\emptyset$, and $\forall x \in$ B $\exists$N$\in N(x)$ N$\cap$A=$\emptyset$.
This follows from choosing N = NB($x$) and lemma 4.2.

This brings us to the central theorem of the paper, the existence of a rule that correctly classifies each $x \in$ A as class_A and each $x \in$ B as class_B.

**Theorem 4.1**

The following classifier correctly classifies $x$ $\forall x \in$ A $\wedge$ $\forall x \in$ B.
    1 if $\exists$N$\in N(x)$ (N$\cap$A=$\emptyset$ $\wedge$ N$\cap$B=$\emptyset$) reject $x$
    2 elseif $\exists$N$\in N(x)$ N$\cap$B=$\emptyset$ classify $x$ as class_A
    3 elseif $\exists$N$\in N(x)$ N$\cap$A=$\emptyset$ classify $x$ as class_B
    4 else reject $x$.

Proof:
Assume $x \in$ A
$\Rightarrow$ $\forall$N$\in N(x)$ $\exists y \in$ N $y \neq x$ $y \in$ A (lemma 4.1)
$\Rightarrow$ N$\cap$A$\neq\emptyset$
$\Rightarrow$ rule 1 does not apply, but $\exists$N$\in N(x)$ N$\cap$B=$\emptyset$ (lemma 4.3)
$\Rightarrow$ rule 2 applies
$\Rightarrow$ $x$ is classified as class_A

Assume $x \in$ B, similar, using rule 3 instead of 2.
$\Rightarrow$ $x$ is classified as class_B

if $x$ is in A rule 4 does not apply as some of its neighbourhoods have just points in A.

q.e.d.

## 5 Remarks

Theorem 4.1 just shows that an error free classifier exists. It does not describe how such a rule may be defined based on a finite set of examples. This will be the topic of the next section. In order to prepare that we have based the theorem not just on the neighbourhood basis, which would have been possible, but formulated it more general in terms of neighbourhood systems.

Rule 1 in theorem 4.1 should take care of that objects not belonging to one of the two classes ($x \notin$ A$\cup$B) are rejected. Some $x \notin$ A$\cup$B, however, having a sufficiently small dissimilarity with objects in A or B may also be classified as A or B and not rejected. This does not contradict the theorem as this considers only points $x \in$ A$\cup$B.

Rule 4 takes care of objects $x \notin A \cup B$ that have small dissimilarities to objects in A as well as B and rejects them.

Further, it can be proven that for each subset S of A the closure of S, cl(S) will be classified as A (objects on the border of A have neighbourhoods that have no points of B and visa versa). So classes are closed sets: they contain their boundary.

# 6 The decision function

We will now proof that the classification function can be expressed in the dissimilarities to a finite set of points if the dissimilarity measure is bounded (any distance is finite) and can be expressed in a finite set of parameters describing the objects. We will set two additional conditions: The following steps are needed:

**Condition 6.1** The number of parameters by which an arbitrary object can be generated is finite, say $n$.

**Condition 6.2** The values of these object parameters are bounded.

**Theorem 6.1** The classifier defined in theorem 4.1 can be based on a finite set of objects.

Sketch of a proof:

1. There exists a one-to-one mapping between the sets of objects, A or B, and $\Re_n$ (condition 6.1)

2. The dissimilarities of a fixed, given object $x$ to all possible objects constitute a continuous mapping $\Re_n \to \Re$ (conditions 3.2 and 6.1).

3. As the object $x$ itself is included in the set of all objects, there is a single point in $\Re_n$ that maps on the origin (condition 3.3).

4. Because of the continuity of the mapping (condition 3.2) a finite interval in $\Re_n$ around $x$ maps on $[0,\varepsilon)$ in $\Re$ and thereby on the neighbourhood NB($x$). Call this interval an object patch.

5. Because the object parameters are bounded (condition 6.2) the entire set of possible objects can be mapped on a finite interval in $\Re_n$. Call this interval for all objects of class A (B) the class patch of class A (B).

6. From 4 and 5 it follows that just a finite set of objects is needed to cover the class patches with object patches.

As a result the classifier can be written as continuous function of a finite set of dissimilarities

**Theorem 6.2** The classifier defined in theorem 4.1 can be written as continuous function of the dissimilarities to a finite set of objects.

Sketch of a proof:

1. Any point can be correctly classified by the nearest neighbour rule applied to the dissimilarities to the set of objects that cover the class patches $P_A$ for class A and $P_B$ for class B (theorem 6.1, step 6).

2. The continuous decision function

   if $\sum_{x \in P_A} \exp(-D((x,y)/s)) - \sum_{x \in P_B} \exp(-D((x,y)/s)) > 0$

   then $y \to$ class_A else $y \to$ class_B

   performs the same classification. It classifies any object correctly if s is sufficiently small, i.e. if $0 < s < (\delta-\varepsilon)/\log(\max(|P_A|,|P_B|))$ as for that value of s the term with the nearest neighbour object dominates.

## 7   Discussion

The given conditions may be met in many practical applications. For instances, if objects can be identified without confusion from a video screen using a finite set of pixels, then for many shape distance measures defined on these pixels the conditions are fulfilled. Consequently, for such applications a zero-error classifier exists (theorem 4.1) and can be based on a finite set of objects (theorem 6.1) by a continuous dissimilarity based classifier (theorem 6.2). As the conditions on the dissimilarity measure are very mild, it does not even have to be a proper metric, such measures may be studied in great freedom using all possible knowledge from the application. It thereby largely enables the inclusion of expert knowledge in statistical pattern recognition procedures.

**Acknowledgements**

This research was supported by the Dutch Organization for Scientific Research (NWO).